# ChatGPT is not A Man but Das Man: Representativeness and Structural Consistency of Silicon Samples Generated by Large Language Models

Dai Li, Linzhuo Li, Huilian Sophie Qiu

**Abstract:** Large language models (LLMs) in the form of chatbots like ChatGPT and Llama are increasingly proposed as "silicon samples" for simulating human opinions. This study examines this notion, arguing that LLMs may misrepresent population-level opinions. We identify two fundamental challenges: a failure in structural consistency, where response accuracy doesn't hold across demographic aggregation levels, and homogenization, an underrepresentation of minority opinions. To investigate these, we prompted ChatGPT (GPT-4) and Meta's Llama 3.1 series (8B, 70B, 405B) with questions on abortion and unauthorized immigration from the American National Election Studies (ANES) 2020. Our findings reveal significant structural inconsistencies and severe homogenization in LLM responses compared to human data. We propose an "accuracy-optimization hypothesis," suggesting homogenization stems from prioritizing modal responses. These issues challenge the validity of using LLMs, especially chatbots AI, as direct substitutes for human survey data, potentially reinforcing stereotypes and misinforming policy.

**Author:** Dai Li, Sociology Dept., China University of Political Science and Law; Linzhuo Li, Sociology Dept., Zhejiang University; Huilian Sophie Qiu, Kellogg school of management., Northwestern University. Corresponding Author: Linzhuo Li, linzhuoli@zju.edu.cn.

## Introduction

Recent developments in Natural Language Processing (NLP) have given rise to human-like chatbots driven by large language models (LLMs), such as OpenAI's ChatGPT and Meta's Llama. These models have garnered significant interest among researchers (Bail 2024; Hase et al. 2021; Liévin et al. 2024; Nazi and Peng 2024; Singhal et al. 2023) and are increasingly considered for their potential to imitate human behaviors, sometimes described as "digital doubles" or "silicon samples" (Argyle et al. 2023; Kozlowski, Kwon, and Evans 2024). However, the present study critically examines this notion, arguing that it may be misleading to conceptualize LLMs as individuals capable of adopting diverse personas or as representative collectives from which reliable, accurate population-level opinions can be retrieved.

Concerns about various biases embedded in LLMs are widespread, ranging from those evident in public discussions to more subtle deviations from human norms (Caliskan, Bryson, and Narayanan 2017; Garg et al. 2018; Potter et al. 2024; Scherrer et al. 2023; Schramowski et al. 2022). For instance, a high-profile case depicted in mass media can be found in the interview of Bassem Youssef by Piers Morgan. Talking about the conflict between Israel and Palestinians in Gaza, Bassem Youssef said the following: "There is a deep sentiment in the Middle East in Arabs that the West do not look at us as equals. So what I did? I went to the machines. I asked ChatGPT simple questions: do Israelis deserve to be free? And you know what they tell me? 'Yes, Israelis deserve the right like any other people.' And then I asked the same question: do Palestinians deserve to be free? And you know what they tell me? 'It is complex; it is a sensitive issue" [1] This anecdote demonstrates that ChatGPT is politically biased.

While some research addresses moral biases or fairness concerns often predicated on specific societal values (Potter et al. 2024; Weidinger et al. 2021), this study focuses on statistical biases related to the representativeness of LLM responses when conditioned on specific demographic profiles. Building on existing inquiries (Bisbee et al. 2024; Boelaert et al. 2025), we identify two fundamental challenges to the validity of LLM-generated synthetic social science data: first, a critical failure to maintain structural consistency, meaning that the proportional accuracy of LLM responses does not hold across different levels of demographic aggregation or intersectional subgroup analysis. Second a pronounced reduction in response variation, a phenomenon we term homogenization, where minority opinions within simulated subgroups are underrepresented.

The issues of structural inconsistency and homogenization are deeply problematic. Functionally, they risk misrepresenting social groups as monolithic entities, potentially reinforcing stereotypes. Methodologically, the lack of realistic variation and internal consistency suggests that using LLM-generated data as direct substitutes for human survey responses is unreliable, likely yielding incorrect sampling distributions for statistical inference. It can even lead to undesired consequences in policy-making. Below we discuss these issues in detail by first developing theoretical dimensions concerning the representativeness of LLM outputs. We propose two hypotheses for the presence of homogenization. We then asked ChatGPT (GPT4) and Meta Llama 3.1 series (8B/70B/405B) two contentious issues on abortion and unauthorized immigrants from the questionnaire of the American National Election Studies (ANES) 2020 Time Series Study [2].

---

[1] Piers Morgan vs Bassem Youssef Round 2 | Two-Hour Special Interview. Accessed at: https://www.youtube.com/watch?v=rqjO5Z9Lt_M, 2024-01-31.

Therefore, in this quickly growing literature, our study builds upon and contributes to existing discussions in primarily two ways. First, we explicitly reveal the problem of representativeness by directly measuring the inherent probabilities of model answers and by directly showing the model's inconsistency across different levels of aggregation, thereby systematically assessing structural consistency. Second, we propose and formalize the accuracy-optimization hypothesis, linking the observed problem of homogenization with the model's algorithmic structure that prioritizes prediction maximization, potentially at the expense of true representativeness. This theoretical framing may help design better statistical workarounds, especially for methods proposed to compare LLM predictions with real observations (Broska, Howes, and Loon 2024).

**Literature Review**

Early explorations into "silicon sampling" demonstrated the potential of LLMs to simulate human-like responses when conditioned on specific demographic profiles.

A prominent and influential example is the work by Argyle et al. (2023), who propose using LLMs like GPT-3 as 'surrogates for human respondents' based on their concept of 'algorithmic fidelity.' Their 'silicon sampling' methodology involves not directly asking model questions (as that would entail great bias in the training data), but conditioning the model on detailed, first-person demographic backstories derived from individual human participants in surveys like the ANES. For each real participant, a corresponding 'silicon subject' is created using their demographic profile (race, gender, age, ideology, partisanship, etc.) as input context. The LLM is then prompted to perform tasks mirroring the original survey, such as predicting presidential vote choice or listing words describing political groups. Crucially, their assessment of fidelity relies on comparing aggregate statistics – like proportion agreement or tetrachoric correlations – calculated within predefined demographic subgroups (e.g., grouped by race, or partisanship) between the silicon sample and the original human data. Although they considered what they term 'algorithmic fidelity,' this approach primarily checks representativeness at some specific, often broader, levels. It at best checks for representativeness along certain dimensions tested (e.g., overall vote choice by party ID).

---
[2] American National Election Studies, ANES 2020 Time Series Study Full Release [dataset and documentation]. 2021: www.electionstudies.org.

Another study by Heyde, Haensch, and Wenz (2024) simulated human voting behavior in Germany using a comparable approach. They created individual "personas" for each of the 1,905 respondents in the 2017 German Longitudinal Election Study (GLES). Each persona was constructed using 13 individual-level variables reported in the survey, including demographics (age, gender, education, and region), economic factors (income and employment), and political attitudes (ideology, party ID, views on immigration, and inequality). These detailed personas were then fed as prompts, in German, to GPT-3.5. The model was tasked with predicting the specific party each simulated individual voted for in the 2017 election. To account for model stochasticity, they generated five vote choice completions for every persona, which were aggregated as the LLM predictions. They then compared LLM predictions to the actual survey data, and found accuracy higher than random.

Building on these methods, subsequent research sought to enhance efficiency. Sun et al. (2024), for example, introduced "random silicon sampling," aiming to emulate human sub-population opinions using only group-level demographic distributions rather than individual-specific data, creating synthetic respondents by randomly assigning characteristics based on a target group's statistical profile.

The practice of generating synthetic responses by conditioning LLMs on specific profiles is not confined to voting behavior simulations. Researchers have also applied similar methodologies across various social science domains (Demszky et al. 2023; Dillion et al. 2023; Kirshner 2024; Sarstedt et al. 2024). For instance, Kozlowski et al. (2024) simulated responses from 2019-era liberals and conservatives to hypothetical pandemic-related questions, aiming to reconstruct the pre-COVID opinion landscape and assess ideological prefiguration, even generating justifications for the simulated opinions. Santurkar et al. (2023) extended this by comparing LLM-generated opinions across numerous topics (from abortion to automation) against those of 60 different US demographic subgroups. Studies have also begun to extend opinion simulations to llm-based agents (Chuang et al. 2023). Going beyond opinion simulation, Veselovsky et al. (2023) used LLMs to generate synthetic text datasets for training classifiers on tasks like sarcasm detection, exploring prompting strategies like grounding with real examples to enhance the "faithfulness" of the generated data for downstream computational social science tasks.

Despite the initial promise and expanding applications, the representativeness of LLM-generated synthetic data has increasingly come under scrutiny. Emerging discussions have begun to probe potential violations of the conditions necessary for valid simulation. Boelaert et al. (2025), for instance, distinguish the "representative hypothesis" (that LLMs can accurately simulate populations) from the "social bias hypothesis" (that failures stem from biased training data). Their findings challenge the representative hypothesis, suggesting LLM's failures are not primarily due to mirroring specific group biases but rather exhibit a distinct "machine bias"—characterized by unpredictable errors and, critically, a substantially lower variance between subpopulations compared to human data. Similarly, Bisbee et al. (2024), while finding that LLM-generated average opinion scores can superficially resemble overall population averages from the ANES, warn that these synthetic data are unreliable for statistical inference. They observe less variation in LLM responses and find that regression coefficients relating respondent characteristics to opinions often differ markedly (sometimes even flipping signs) compared to estimates using real survey data.

These critiques make valuable contributions by empirically demonstrating discrepancies and raising important cautionary flags. However, their focus tends to highlight symptoms of potentially deeper representational issues—such as compressed variance or specific mismatches at aggregated levels—often mixed with other sources like prompt sensitivity or temperature change. While hinting at underlying problems, particularly Bisbee et al. (2024)'s findings on differing correlational structures, these studies often leave the fundamental challenges to faithful representation underexplored.

In addition, early studies sometimes only focus on pre-trained models (laboratory-based LLMs), yet in social scenarios LLMs are often further tuned to be chatbots to facilitate interaction with human, mostly using reinforcement learning based on human preferences (Ouyang et al. 2022; Schulman et al. 2017) or Direct

Preference Optimization (DPO, Rafailov et al. 2024). This tuning process often involves the implementation of "guardrails" by developers, designed to prevent chatbots from generating responses on highly controversial, inflammatory, or otherwise deemed unacceptable topics (Ayyamperumal and Ge 2024; Yang et al. 2024). These guardrails are themselves shaped by specific cultural and political contexts. Consequently, the presence and nature of these guardrails can significantly impact the LLM's output when simulating opinions on a wide range of social and political issues, potentially leading to an underrepresentation or distortion of views that might be prevalent in actual human populations but are filtered out by the model's safety mechanisms. This introduces another layer of complexity when assessing the "algorithmic fidelity" of such models, as the simulated opinions may reflect the persona's characteristics not just in a vacuum but within the content moderation policies embedded within the LLM. Evaluating these chatbot LLMs, therefore, is more crucial and requires consideration of these inherent limitations.

Building upon these insights, we propose that a robust evaluation of LLM-generated synthetic data requires a focus on at least two critical dimensions of representativeness: structural consistency in accurately mirroring distributions across all levels of aggregation, and the preservation of realistic inter-subject variation.

**Theoretical Framework**

**The Challenge of Structural Consistency**

A crucial prerequisite for leveraging LLMs to generate valid synthetic social science data is that the underlying ground-truth distribution of the model's knowledge for a chosen target population must be accurate. This means the model must contain "knowledge/preference distributions" that, when sampled under specific conditions, proportionally mirror the real-world distribution of characteristics, attitudes, or behaviors within the targeted population subgroup. Simply put, if a particular demographic group in reality exhibits a specific split in opinions or actions (e.g., 80% hold belief A, 20% hold belief B), an LLM should, when conditioned on that group's profile, generate outputs reflecting this same 80/20 proportion across numerous simulations.

Building on this, we identify a further critical property: Structural Consistency. This property requires the model to maintain proportional accuracy consistently across all possible levels of aggregation, from the coarsest single-variable subgroups (like gender or race) down to the most highly specific, intersectional subgroups defined by multiple demographic variables simultaneously (e.g., simulating the precise opinion distribution among 20-30 ages, low-income, rural, Black men without a college degree). Achieving structural consistency implies that the model possesses a property akin to "closure under aggregation" regarding representativeness; the accuracy of proportional mirroring holds consistently regardless of how finely or broadly the subgroups are defined.

An LLM might demonstrate a relatively high accuracy from a specific level of granularity towards aggregated statistics (e.g., correctly simulating the overall vote split between men and women when aggregated solely by gender) without necessarily possessing Structural Consistency. This superficial accuracy at one level provides no guarantee about representativeness at finer granularities (e.g., young men vs. old men) or if different paths of aggregations are considered. Therefore, for LLMs to truly reliably simulate human samples, structural consistency should be satisfied. Previous studies advocating the value of LLM simulation, including the foundational work by Argyle et al. (2023) and Heyde et al. (2024), however, often overlook the stringent demands implied by this property.

**Understanding Reduced Variation and Response Homogenization**

Beyond the challenge of structural consistency, another significant concern, underscored by the findings of Boelaert et al. (2025) and Bisbee et al. (2024), is the observed reduction in response variance, termed response homogenization, within LLM-generated samples compared to human data. While this phenomenon has been noted, systematic explanations for its origins within the context of LLM-based social simulation remain relatively underdeveloped. We consider two primary hypotheses that might account for this observed homogenization.

The first, which we term the Biased Sample Hypothesis, assumes that an LLM primarily retrieves information from an innate stock of knowledge representing the world. Homogenization, from this viewpoint, arises because this internal knowledge base is itself biased. Presumably, the vast quantities of text data used for LLM training do not offer a statistically accurate reflection of actual human opinion distributions. Due to various socio-economic factors, not all individuals or groups have equal opportunities to voice their opinions in forums that contribute to training corpora, nor do all opinions have an equal chance of propagation. Such selection biases could lead to the overrepresentation of certain viewpoints and the underrepresentation of others within the training data, resulting in an LLM that internally embodies a skewed sample. This bias is often recognized as the major problem (Bender et al. 2021; O'neil 2017).

An alternative perspective, which we refer to as the Accuracy Optimization Hypothesis, suggests that response homogenization is not merely a passive reflection of a biased internal sample but a consequence of the LLM's training objectives. LLMs are generally trained to maximize the accuracy or likelihood of their outputs given an input (Cho, Kim, and Kim 2024; Ge et al. 2023; Li et al. 2023; Rybakov et al. 2024). When tasked with predicting opinions for a user or persona whose true beliefs are unknown, the most optimized strategy, as we demonstrate formally below, is often to respond with the mode, or most prevalent, opinion known to the LLM [3].

Suppose on a certain issue, there exists a finite number of beliefs, $x_1, \cdot, x_i, \cdot, x_n$. Proportion $p_i$ of the population believes in a certain belief, $x_i$. When answering a question about this issue by a user whose belief is unknown to the model, the model answers with belief $x_i$ with a probability of $q_i$. If the model's answer matches the user's belief, then the model has made a successful prediction. The model is trained to maximize the chance of success. Under these conditions, what is the most optimized strategy?

For starters, consider that there are only two beliefs, $x_1$ and $x_2$. If the model answers with $x_1$ with $q_1$ probability, the chance of it matching the user's belief is $r = q_1 p_1 + (1 - q_1)(1 - p_1)$. The derivative of this function is $\frac{dr}{dq_1} = p_1 - (1 - p_1) = 2p_1 - 1$. If $p_1 > 0.5$, $2p_1 - 1 > 0$. This means that as long as belief $x_1$ is more favored among the population, the best strategy of the model is to always answer with $x_1$.

This can be generalized to where there are $n$ beliefs about a given issue. Suppose there are a finite number of beliefs, $x_1, \ldots, x_n$, $\sum(p_1, \ldots, p_n) = 1$, and $p_1 > p_2 > \cdot > p_n$. The expectation of user satisfaction is $U(q) = \sum_i p_i * q_i$. Then we have

$$U <= \sum p_1 * q_i = p_1 \sum q_i = p_1$$

---

[3] There is a prolific vein of research on the trade-off between accuracy and diversity of predictions (see Peng et al. 2023), which is relevant to our present problem.

Therefore, the maximum of $U$ is obtained when $q_1 = 1, q_2 = \cdot = q_n$, which means to always answer with the mode.[4] Notably, this hypothesis is not predicated on the premise that LLM retrieves information from a structured sample. Its innate stock of knowledge can be self-conflicting and without consistency.

This hypothesis suggests that asking an LLM to respond from a persona is not akin to sampling an individual from a population who holds genuine, potentially idiosyncratic opinions. Instead, the LLM, optimized for accuracy, is incentivized to generate responses that align with the most broadly probable or dominant views attributed to the subgroup of that persona, leading to homogenization irrespective of the true diversity within its internal knowledge.

Philosophically speaking, the accuracy optimization hypothesis suggests that an LLM is possibly the ultimate 'Das Man' (the They, Heidegger 1962) rather than 'a man.' This notion of 'Das Man' is used to describe the 'crowd,' the 'herd,' the unreflective thinking of tradition or masses. This hypothesis is not predicated on the LLM possessing an internally coherent or consistently structured knowledge sample; its innate knowledge could be fragmented or even self-contradictory without impeding its ability to identify and reproduce modal responses for perceived accuracy.

Evaluating these two hypotheses, we find the Accuracy Optimization Hypothesis to be a probable but neglected possibility for the observed response homogenization, particularly as it goes beyond the issue of data representation. While selection biases in training data (as per the Biased Sample Hypothesis) might influence the LLM's perception of which opinion is modal, the optimization process itself would still drive the model to output that perceived mode with high probability, thereby contributing significantly to homogenization.

---

[4] We thank Qingcan Wang for helping us with the proof.

**Data and Method**

**ANES 2020 data**

We use ANES 2020 Time Series Study data to estimate political opinions held by non-institutional U.S. citizens aged 18 or older living in the 50 U.S. states or the District of Columbia. We specifically focus on two questions from the questionnaire:

V201336: There has been some discussion about abortion during recent years. Which one of the opinions on this page best agrees with your view? You can just tell me the number of the opinion you choose.

1. By law, abortion should never be permitted.

2. The law should permit abortion only in case of rape, incest, or when the woman's life is in danger.

3. The law should permit abortion for reasons other than rape, incest, or danger to the woman's life, but only after the need for the abortion has been clearly established.

4. By law, a woman should always be able to obtain an abortion as a matter of personal choice.

5. Other (Specify)

V201417: Which comes closest to your view about what government policy should be toward unauthorized immigrants now living in the United States? You can just tell me the number of your choice.

1. Make all unauthorized immigrants felons and send them back to their home country.

2. Have a guest worker program that allows unauthorized immigrants to remain in the United States in order to work, but only for a limited amount of time.

3. Allow unauthorized immigrants to remain in the United States and eventually qualify for U.S. citizenship, but only if they meet certain requirements like paying back taxes and fines, learning English, and passing background checks.

4. Allow unauthorized immigrants to remain in the United States and eventually qualify for U.S. citizenship, without penalties.

We focus on these two questions for methodological reasons. These two questions are expected to be well suited for an LLM to answer because they are multiple-choice questions whose choice options are substantive and distinct. LLMs are commonly trained and assessed when the question is presented in an explicit multiple-choice format (Robinson 2022) and the model only needs to answer with the item number. In addition, in recent years, reproductive rights and unauthorized immigration have become two of the most high-profile political issues in the United States. Presumably, there are plenty of relevant texts for LLMs to train on. Studying these two questions best connects our research with the current political reality. To test for robustness of our finding, we provide some auxiliary analyses on other questions in the Supporting Information. These analyses confirm that multiple-choice questions are suitable for LLMs and our findings are robust.

We control for the following covariates: sex (V201600), race (V201549x), education (V201511x), and religion (V201435). These variables are all from the pre-election dataset, so they are weighed by full sample pre-election weight (V200010a). We use these covariates because they are significantly correlated to the opinions given by interviewees, based on two multinomial logistic regressions we ran (See our codes upon publication). Other covariates in consideration include: age, sex orientation, marital status, family income, and employment status. Since our purpose in this study is not to find the best models that explain the formation of political opinions, we decide to only focus on the four aforementioned covariates (sex, race, education, and religion) for simplicity.

Table1: Descriptive statistics of sex, race, education, religion, and answers to questions on abortion and immigration in ANES 2020 data. Unweighted.

| Variable | Id | Level | Count | Percentage |
|---|---|---|---|---|
| sex | S1 | Male | 3,101 | 47.89 |
| | S2 | Female | 3,374 | 52.11 |
| race | R1 | White, non-Hispanic | 4,936 | 76.23 |
| | R2 | Hispanic | 614 | 9.48 |
| | R3 | Black, non-Hispanic | 362 | 5.59 |
| | R4 | Asian or Native Hawaiian/other Pacific Islander, non-Hispanic alone | 265 | 4.09 |
| | R5 | Multiple races, non-Hispanic | 188 | 2.90 |

| Variable | Id | Level | Count | Percentage |
|---|---|---|---|---|
| | R6 | Native American/Alaska Native or other race, non-Hispanic alone | 110 | 1.70 |
| education | E1 | Less than high school credential | 275 | 4.25 |
| | E2 | High school credential | 973 | 15.03 |
| | E3 | Some post-high school, no bachelor's degree | 2,118 | 32.71 |
| | E4 | Bachelor's degree | 1,754 | 27.09 |
| | E5 | Graduate degree | 1,355 | 20.93 |
| religion | relgA | Protestant | 2,061 | 31.83 |
| | relgB | Roman Catholic | 1,600 | 24.71 |
| | relgC | Nothing in particular | 1,420 | 21.93 |
| | relgD | Agnostic | 452 | 6.98 |
| | relgE | Atheist | 333 | 5.14 |
| | relgF | Orthodox Christian (such as Greek or Russian Orthodox) | 153 | 2.36 |
| | relgG | Jewish | 185 | 2.86 |
| | relgH | Latter-Day Saints (LDS) | 113 | 1.75 |
| | relgI | Muslim | 46 | 0.71 |
| | relgJ | Buddhist | 72 | 1.11 |
| | relgK | Hindu | 40 | 0.62 |
| | relgL | Something else | 0 | 0.00 |
| abortion | abortion1 | 1 | 627 | 9.68 |
| | abortion2 | 2 | 1,463 | 22.59 |
| | abortion3 | 3 | 890 | 13.75 |
| | abortion4 | 4 | 3,228 | 49.85 |
| | abortion5 | 5 | 235 | 3.63 |
| | Missing | | 32 | 0.49 |
| immigration | immigration1 | 1 | 800 | 12.36 |
| | immigration2 | 2 | 958 | 14.80 |
| | immigration3 | 3 | 3,549 | 54.81 |
| | immigration4 | 4 | 1,131 | 17.47 |
| | Missing | | 37 | 0.57 |

Table 1 presents the levels of these four categorical covariates. Theoretically, they can make 720 combinations that describe different subgroups of people. We first eliminate "Something else" from religion, because it lacks information with the absence of other choices when fed to an LLM. Second, some subgroups of people are not represented in the survey data. The final number of subgroups we arrive at is 395.

**Models**

The primary LLMs we study are ChatGPT's gpt-4-613 (subsequently GPT4) and Meta's Llama 3.1 series 8B, 70B, and 405B (subsequently Llama8B/70B/405B). ChatGPT's gpt-4-613 has a knowledge cutoff date of Dec 01, 2023, while the Llama models have a cutoff date of Dec 2023. [5] This means that these models are qualified to answer questions about the world before these dates.

The GPT4 API has three types of input: system message, assistant message and user message. In this research, assistant message is not used; system message is used to assign the model a persona to produce a 'digital double' of a person in a subgroup; and user message is the query. With the API, GPT4 provides output in at least two ways. It can output the mere response or generate response tokens and their log probabilities using the parameter *logprobs*.[6] If this parameter is included via openAI's API, GPT4 will return the log probabilities of token predictions along with the output tokens, and because the log probabilities are inherent to the model's underlying computing process, they would stay the same every time it is queried. Notably, the log probabilities reported via API reflect the model's raw output distributions before any sampling strategy is applied, and thus remain independent of the temperature parameter setting and are particularly useful for our research on subgroup response patterns. This independence is methodologically valuable as it allows us to access the model's genuine distribution in its predictions regardless of the randomness introduced during generation. The log probabilities can later be transformed into probabilities of answers given by a subgroup. In this study, we set the parameter, *logprobs*, to be 5-10 to match the number of options.

---

[5] See https://platform.openai.com/docs/models/gpt-4 and https://docsbot.ai/models/llama-3-1-405b-instruct. Accessed on April 30, 2025.

[6] See https://platform.openai.com/docs/api-reference/chat/create. Accessed on April 30, 2025.

For the Llama series, we employed all three sizes of the instruction-tuned Llama 3.1 models: 8B, 70B, and 405B. These models are also decoder-only transformer models, fine-tuned for instruction following using supervised fine-tuning and direct preference optimization. While GPT-4's model structure is not fully disclosed, Llama 3.1 models are open-source, and their structures are publicly available. This openness allows for potential robustness checks of our results. We accessed these models via the Together AI API [7].

**Prompts**

For GPT4, the prompts are constructed with the following scheme. The system message is used to assign a persona to the model. The message is constructed as: "You are a US citizen with voting rights. It is the year 2020. You are responding to a survey. You are {sex}. You are {race}. Your highest education is {education}. Your religious identity is {religion}." The actual values of sex, race, education and religion in ANES 2020 are inserted in these corresponding places. Since there are 395 distinct combinations of these variables in the data, there are 395 distinct system messages for 395 personas. For example, for the combination of (sex = male, race = Hispanic, education = Graduate degree, religion = Roman Catholic), the system message is:

"You are a US citizen with voting rights. It is the year 2020. You are responding to a survey. You are male. You are Hispanic. Your highest education is a Graduate degree. Your religious identity is Roman Catholic."

The user message is constructed using the questions from the questionnaire of ANES 2020, asked almost verbatim. For example, the message on abortion is:

"There has been some discussion about abortion during recent years. Which one of the opinions on this page best agrees with your view? Just tell me the number of the opinions you choose. 1. By law, abortion should never be permitted. 2. The law should permit abortion only in case of rape, incest, or when the woman's life is in danger. 3. The law should permit abortion for reasons other than rape, incest, or danger to the woman's life, but only after the need for the abortion has been clearly established. 4. By law, a woman should always be able to obtain an abortion as a matter of personal choice. 5. Other."

---

[7] https://api.together.xyz/

We intentionally keep the format of the question so that if the distribution of responses is different from that in the survey, the reason is unlikely to be that the questions are posed differently.

There are 395 by 2 combinations of system message and user message. We ask each of them once independently via API. In response, we get 395 by 2 responses from GPT4.

For the Llama 3.1 models (8B, 70B, and 405B), accessed via the Together AI API, a similar approach to persona assignment and question-posing was used, but the method for obtaining probabilities for each answer choice differed due to API functionalities and model interaction paradigms. We adapted the prompt format based on templates suggested by Meta for user dialogues, and implemented a prompt strategy designed to give us complete control over the specific output sequence whose likelihood was being determined. For each survey question and each of the 395 defined personas, we queried the model five times—once for each possible answer option. The prompt structure was formatted as follows: messages=[{"role": "user", "content": "{persona_and_question_text}"}, {"role": "assistant", "content": "The answer would be {N}"}]

In this structure: {persona_and_question_text} contains the full persona declaration followed by the survey question and its enumerated options. {N} represents the specific answer choice (1, 2, 3, 4, or 5) whose probability we aimed to assess. For each of these five queries per persona and question, we obtained the log probabilities of the tokens that constitute the assistant's specified response (i.e., "The answer would be {N}"). By summing these token log probabilities for the constrained assistant message and then converting this sum to a probability, we derived a score for that specific answer choice. These five scores were then normalized to sum to 1, yielding a probability distribution over the five choices for that particular persona and question.

Despite slightly different prompting methods for closed and open-sourced models, the results concerning the problems of silicon samples are highly consistent.

## Measures

### Model accuracy

Before detailing our probability-based evaluation method, it is important to clarify its purpose and limitations. While not the only way to evaluate LLM's outputs, we employ this approach primarily to examine the potential for LLMs to produce overly homogenized and inconsistent representations of demographic groups, thereby reinforcing misconceptions about between-group differences and within-group homogeneity. Our probability-based measurement thus deliberately extends methodologies from previous silicon sampling studies (Argyle et al. 2023; Bisbee et al. 2024; Heyde et al. 2024; Sun et al. 2024), not because we endorse the silicon sample approaches or believe in the comparability between LLM outputs and human samples, but because it provides a relatively consistent framework for examining representation issues. In addition, our measure also has its advantages: it efficiently calculates expected match rates without requiring extensive sampling in quantifying distributional differences between human and silicon samples, and it remains relatively independent of model temperature and other hyper-parameters that might otherwise confound our analysis. That said, we fully acknowledge the inherent challenges in directly comparing algorithmic outputs with human survey responses. Even with its limitations, however, we show that our method is able to reveal the significant problems of silicon samples.

Specifically, we evaluate the performance of LLM's silicon samples in two dimensions: accuracy and variation. The former is a routine dimension of performance of machine learning algorithms for classification tasks. The latter is a separate dimension that we aim to bring to the dialogue about the political and social issues of LLMs.

For each persona group $i$, accuracy, $U_i$, is measured as the dot product (also can be understood as the cosine similarity) between the model's predicted probability distribution over response options and the actual distribution of responses for that demographic group in the ANES survey data.

$U_i = \sum_{j=1}^{5} p_{ij} q_{ij}$, where $q_{ij}$ refers to the model's predicted proportion of choosing response $j, j = 1,2,3,4,5$ under persona $i$, and $p_{ij}$ is the actual proportion calculated from the ANES data.

Intuitively, $U_i$ measures how accurately models capture the real opinion distribution within each demographic group $i$. This calculation method is theoretically equivalent to the more computationally intensive one-to-one sampling approach: if we prompted the model to generate $n_i$ responses for demographic group $i$ (matching the actual subgroup population size), and each time allowed the model to generate its answer according to its predicted probability distribution over options (a multinomial distribution), the resulting proportion of responses would converge to the model's predicted response distribution as $n_i$ becomes large.

$U_i$ thus efficiently calculates the match rate between model-generated samples and the true population distribution without requiring extensive sampling procedures that were common in previous research (Argyle et al. 2023; Bisbee et al. 2024; Heyde et al. 2024).

We then evaluate accuracy across six settings. Four are LLMs: GPT4, Llama8B/70B/405B. Two serve as theoretical benchmarks: (1) The 'ANES self-similarity' benchmark, calculated as the sum of squared probabilities of the original ANES distribution ($U_i = \sum_{j=1}^{5} p_{ij}^2$), which measures the inherent concentration of opinions within each demographic group; and (2) The 'Mode' benchmark, which always predicts the most common response for each persona, yielding accuracy $U_i = p_m$ (where $p_m$ is the probability of the most frequent response). These benchmarks provide good references: since perfect accuracy is mathematically impossible due to the inherent diversity of opinions within demographic groups, we evaluate LLMs relative to these optimized baselines rather than against an absolute 100% scale. The Mode benchmark – always predicting the most common answer – actually achieves the theoretically highest possible expected accuracy for any model attempting to predict group-level distributions, as proved before. This is intuitive: if you must assign your entire 'probability budget' to one option under a given persona, you should place it on the option that occurs most frequently in the true distribution. Any other allocation would dilute your prediction toward less common outcomes, reducing the expected match rate. Figure 1 below illustrates how accuracy is calculated across the four models and two benchmarks.

# Theoretical Equivalence to One-to-One Silicon Sampling

**Example Persona**

"You are a US citizen with voting rights. It is year 2020. You are responding to a survey. You are female. You are Hispanic. Your highest education is college graduate. There has been some discussion about abortion during recent years. Which one of the opinions best agrees with your view? 1..2..3..4..5.."

**Calculating Accuracy**

$$U_i = \sum_{j=1}^{5} p_{ij} \times q_{ij}$$

**Actual ANES Survey Distribution**

Real-world opinion distribution for the given persona group

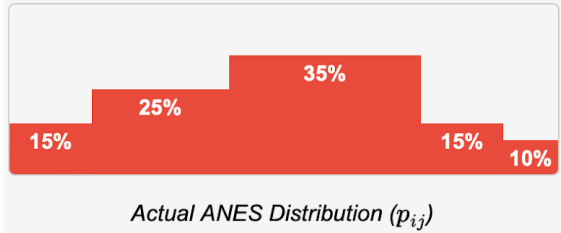

Actual ANES Distribution ($p_{ij}$)

**LLM Predicted Distribution**

Model predicts probabilities for each response option (1-5 scale)

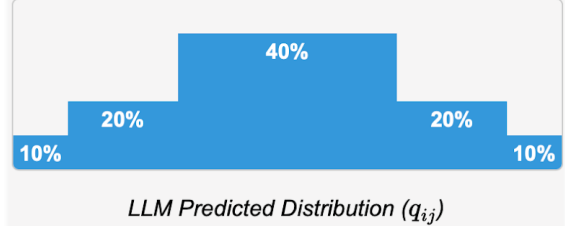

LLM Predicted Distribution ($q_{ij}$)

**Benchmarking Against Theoretical Optimum**

**Benchmark: ANES (Self-Similarity)**

Sum of squared probabilities

$$U_i = \sum p_{ij}^2 = 0.26$$

*Inherent concentration measure*

**Benchmark: Answer with Mode**

Always predicting most common response

$$U_i = p_{\text{mode}} = 0.35$$

*Theoretical maximum accuracy*

*Figure 1* Illustration of probability-based assessment of accuracy.

**Structural consistency**

We measure whether the LLMs have structural consistency based on the basic assumption often overlooked in traditional survey research. For a real survey, it is obvious that statistics of subgroups at lower levels can always be aggregated to statistics of subgroups at higher levels. For example, the proportions of choosing '1' for men and women (sex) can be aggregated from the proportions of choosing '1' for black men, black women, non-black men, and non-black women (sex × race). This is because they all come from the individual observation in the sample.

We can test this for LLMs, similarly, to see whether the size of each subgroup can be aggregated to higher levels. For example, when asked to estimate the number of men, if the LLM responds independently that (1) there are 100 black men, (2) 102 non-black men, and (3) there are 300 men in total, the aggregation (202 men) would not match the response (300 men), violating the consistency assumption.

To do so, first, we construct subgroups at different levels: zero (population), one (sex), two (sex × race), three (sex × race × education), and four (sex × race × education × religion). Second, we assign these subgroups to the model and ask the two questions. Third, we aggregate lower level (more covariates combined) statistics to predict higher level statistics using ANES weights according to the silicon sample approaches in previous studies.

For example, suppose the distribution of probabilities of an LLM answering '1-5' to the abortion question is $D_1$ for personas constrained on one variable (say, sex), and $D_2$ for personas on two variables (say, sex and religion). We can then aggregate religion subgroup answers to sex $D_1^2$, and compare $D_1$ with $D_1^2$ to see whether the two distributions match. Note that if more variables are included, then structural consistency remains an easy task for real surveys but may become more difficult for LLMs. We thus focus only on at most four demographic variables, though the consistency issues we identify would likely intensify with additional variables.

**Model variation**

Compared to consistency, the measure of variation is relatively straightforward. Variation is measured by variation ratio, which calculates the percentage of non-mode results. Given $p_m$ as the proportion of the mode (the most frequently chosen option in a dataset), the variation ratio is simply $1 - p_m$.

**Results**

**Prediction accuracy by fine-grained subgroups**

Figure 2 shows the smoothed distributions of accuracy across all 395 personas for different models. GPT4, Llama405B, and Llama70B show accuracy distributions between 0.50 and 0.60, with left tails indicating personas with lower accuracy. Llama8B clearly underperforms compared to the other LLMs. Compared to baselines (always selecting the most common response or directly using ANES distribution data), LLMs exhibit fewer personas with high accuracy scores (as shown by the shorter right tail) and more personas with low accuracy scores (evident in the extended left tail).

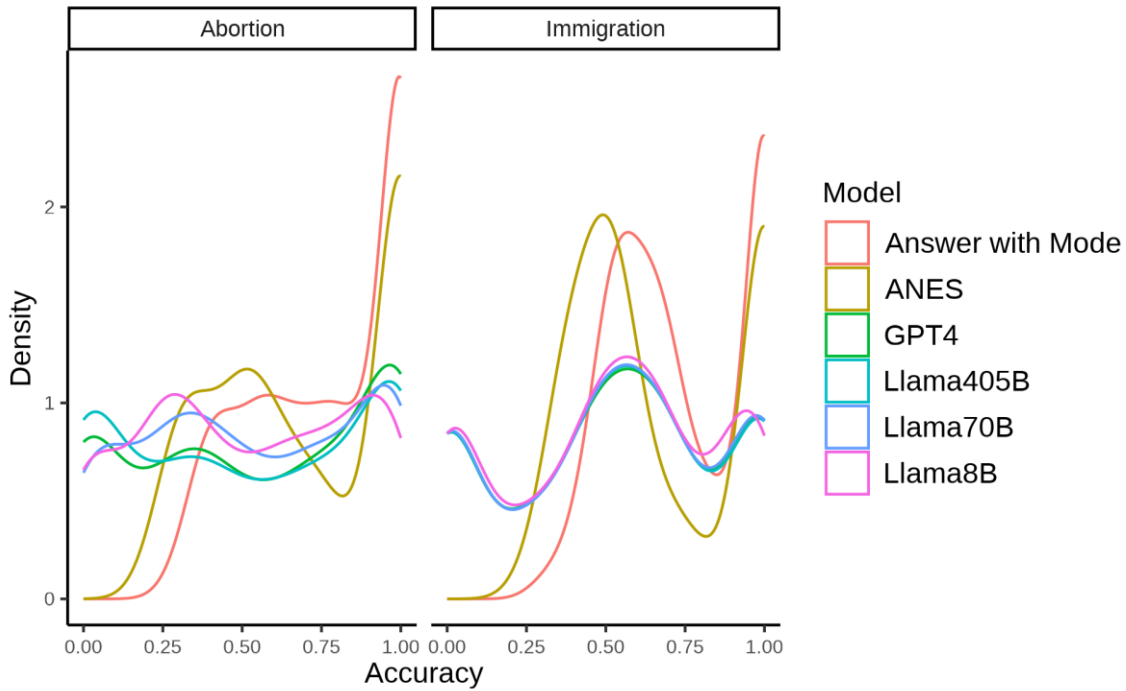

*Figure 2* Distribution of accuracy for Abortion and Immigration by persona subgroups for all models and benchmarks.

Table 2 shows the mean accuracy of all models and benchmarks. The reported mean accuracies are either unweighted (every subgroup having the same weight) or weighted by ANES 2020 pre-election weights. The difference between weighted and unweighted indicates how accuracy is uneven among subgroups of different population sizes. For the abortion question, the 'Answer with Mode' strategy yields an unweighted accuracy of 0.77 and a weighted accuracy of 0.56. The ANES data has an unweighted accuracy of 0.70 and weighted accuracy of 0.44. GPT4 achieves an unweighted accuracy of 0.55 and weighted accuracy of 0.49. Llama405B shows 0.52 (unweighted) and 0.47 (weighted), Llama70B shows 0.53 (unweighted) and 0.48 (weighted), and Llama8B performs the lowest among LLMs with 0.51 (unweighted) and 0.44 (weighted). For the immigration question, "Answer with Mode" achieves 0.74 (unweighted) and 0.59 (weighted). ANES data shows 0.67 (unweighted) and 0.45 (weighted). GPT4, Llama405B, Llama70B, and Llama8B all achieve similar weighted and unweighted accuracies. The results confirm that the most accuracy-optimized strategy is to always answer with the mode (Red line). In reference to the maximum expected accuracy (0.77 for abortion and 0.74 for immigration) offered by this strategy, the LLMs' accuracy performance is moderately good (reaching around 70% of the maximum). The smallest model Llama 3.1 8B performs a bit lower, suggesting that the scaling up model sizes may slightly help with accuracy.

Table2: Mean model accuracies regarding questions on abortion and immigration, weighted and unweighted.

| Model | Question | Unweighted Accuracy | Weighted Accuracy |
|---|---|---|---|
| Answer with Mode | Abortion | 0.77 | 0.56 |
| ANES | Abortion | 0.70 | 0.44 |
| GPT4 | Abortion | 0.55 | 0.49 |
| Llama405B | Abortion | 0.52 | 0.47 |
| Llama70B | Abortion | 0.53 | 0.48 |
| Llama8B | Abortion | 0.51 | 0.44 |
| Answer with Mode | Immigration | 0.74 | 0.59 |
| ANES | Immigration | 0.67 | 0.45 |
| GPT4 | Immigration | 0.53 | 0.55 |
| Llama405B | Immigration | 0.53 | 0.55 |
| Llama70B | Immigration | 0.53 | 0.55 |
| Llama8B | Immigration | 0.52 | 0.55 |

**Structural inconsistency**

Figure 3 and 4 show how silicon samples generated by GPT4 and Llama series may violate structural consistency. For each figure, we computed the predicted probability distribution among options (see the method section on "prompts" for details) using demographic variables at different levels of granularity. '0' means we directly ask the model to answer questions without persona setting. '1 Vars' refers to persona by sex, '2 Vars' refers to persona by sex and race, '3 Vars' by sex, race and education, and '4 Vars' by sex, race, education and religion. The abortion topic clearly shows a significant mismatch among different persona settings for all models inspected. Particularly, the mismatch is largest for 4 Vars. For instance, for GPT4 and Llama 405B, the dominant answer for "4 Vars" is option 2: "The law should permit abortion only in case of rape, incest, or when the woman's life is in danger", where for the rest categories the dominant answer becomes option 4: "By law, a woman should always be able to obtain an abortion as a matter of personal choice." Llama 70B are slightly more consistent, but the distribution still differs a lot from 4 Vars to 0 Vars. Llama 8B shows more diversity among all answers, and still has quite significant changes in distributions. LLM simulations under the immigration topic show less discrepancies for all models, we will show later that this is likely because of another problem – homogeneity of answers. LLM answers for the immigration topic displays far more severe homogeneity than for the abortion topic.

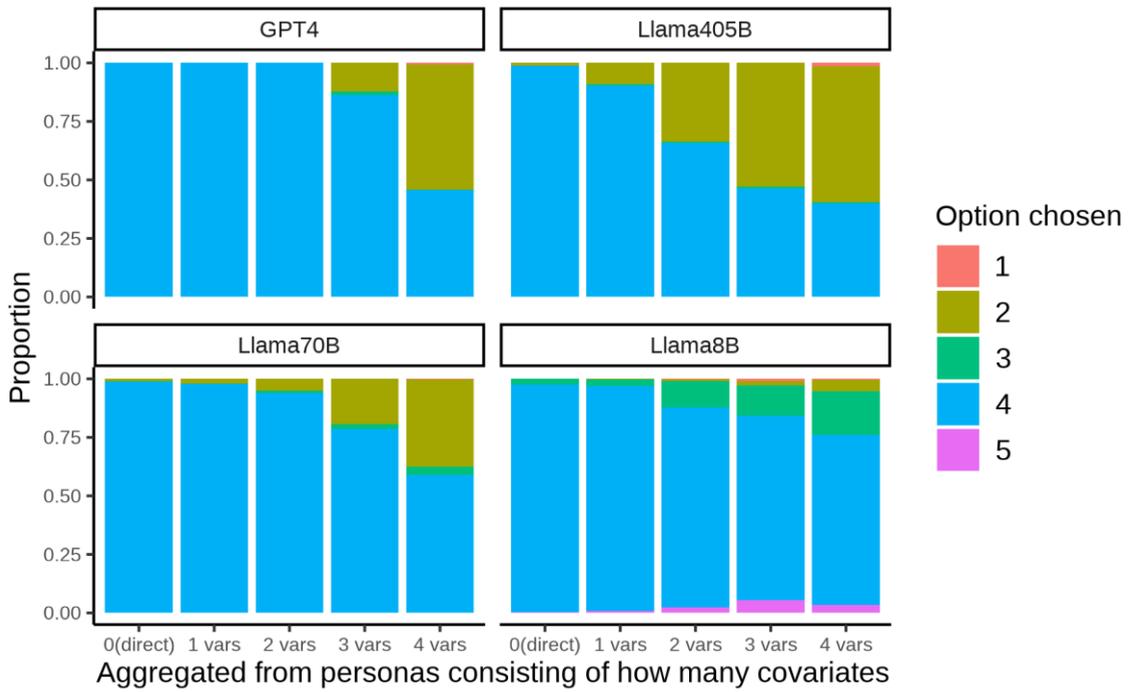

***Figure 3*** *Predicted probability distribution of options for the abortion topic under various persona granularity.*

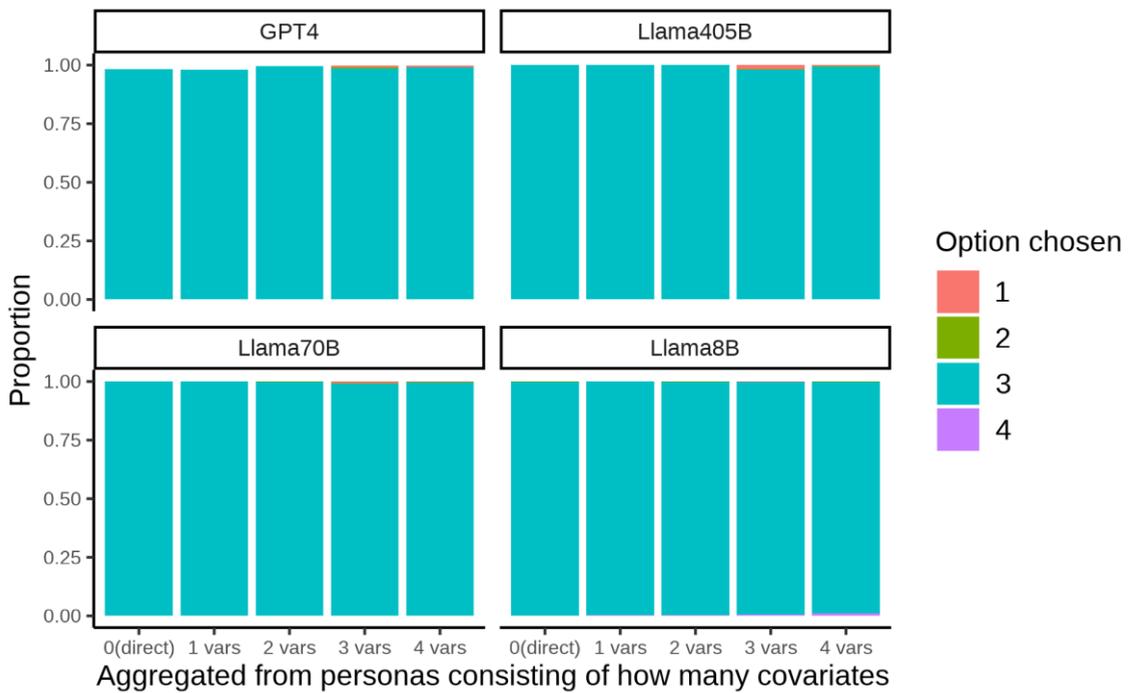

***Figure 4*** *Predicted probability distribution of options for the immigration topic under various persona granularity.*

Structural consistency would also entail that statistics (like accuracy) derived from finer-grained personas, when aggregated to a coarser level, should also match the statistics derived from directly querying personas at that coarser level. Figure 5 illustrates a test of this: it displays accuracy statistics where results from personas constructed using two, three, or four demographic variables were first obtained, and then these results were aggregated to the 1-variable level (sex). These aggregated accuracies are compared with the accuracy from direct one-variable (sex-only) persona queries. If LLMs were structurally consistent, the accuracy lines for '1 Variable' (direct query with sex), '2 Variables' (aggregated from sex and race), '3 Variables' (aggregated from sex, race and education), and "4 Variables" (aggregated from sex, race, education and religion) should largely overlap for each model and sex category on the radar plot.

Observing the abortion topic (Figure 5a), a clear lack of structural consistency is evident across all tested LLMs. For example, for GPT4 (female personas), the accuracy obtained from direct 1-variable queries is visibly higher (the light purple line, ~0.5) than the accuracy obtained when results from 4-variable personas are aggregated to the 1-variable (sex) level (the darkest purple line, ~0.35). This means the accuracy of representing female opinion on abortion changes depending on whether the model is queried with a simple "female" persona or if a more detailed persona's output (e.g., "female, specific race, specific education, specific religion") is aggregated back to the "female" level. Similar divergences between the differently aggregated accuracy lines are observable for Llama 405B, Llama 70B, and Llama 8B, for both male and female categories. The distinct separation of these lines for any given model indicates that the level of persona granularity used in the underlying query significantly impacts the resulting aggregated accuracy at the base 'sex' variable level, a hallmark of structural inconsistency.

For the immigration topic (Figure 4b), the aggregated accuracy lines for different levels of persona granularity appear more closely clustered for some models, suggesting potentially greater stability than observed for the abortion topic. However, deviations indicative of imperfect structural consistency are still present. For instance, with GPT4, the accuracy from direct 1-variable queries (lightest purple line) for both male and female personas tends to be slightly different from the accuracies derived from aggregating 2, 3, or 4-variable persona results. In the scenario with real survey data, the results would align. Thus, the observed discrepancies, even if less pronounced for immigration than for abortion, still point towards LLMs not adhering to this principle of structural consistency.

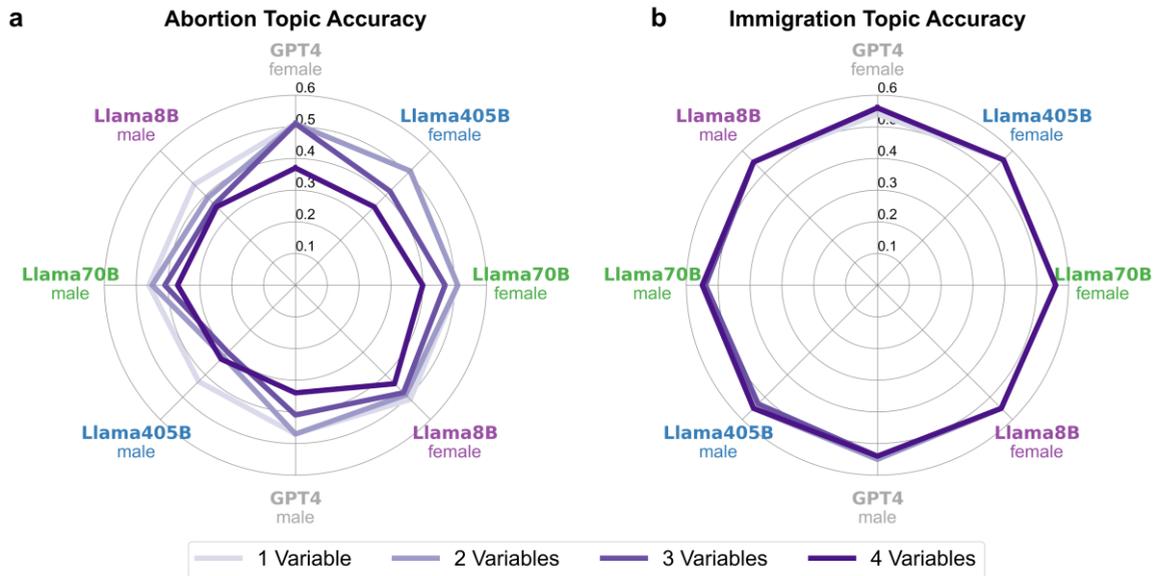

*Figure 5* Structural Inconsistency of Silicon Samples.

**Homogeneity in Silicon Samples**

A second key finding of our study is the significant homogenization of responses from LLMs. We measure this using the variation ratio (VR), where a lower VR indicates higher homogenization (i.e., the predominance of a single answer). Figures 6 and 7 display heatmaps of variation ratios across numerous demographic subgroups for the abortion and immigration questions, respectively, comparing real ANES data with GPT4 and the Llama series models outputs.

For the abortion question (Figure 6), the ANES data (panel a) shows a diverse range of VRs across subgroups, with a P(VR<0.05) of 0.4, meaning 40% of subgroups have a dominant opinion held by over 95% of the subgroup. In stark contrast, GPT4 (panel b) and Llama405B (panel c) exhibit much lower VRs across most subgroups, with P(VR<0.05) at 0.8 for both—indicating extreme homogenization in 80% of subgroups. While Llama70B (panel d) shows a P(VR<0.05) of 0.4, and Llama8B (panel e) shows a P(VR<0.05) of 0.3, similar to ANES, we should note that VR<0.05 is only a threshold value set arbitrarily. Comparing the overall distributions of VR (or by looking at the brightness of their heatmaps) still shows that LLMs have higher homogenization than ANES.

A similar, even more pronounced pattern is evident for the immigration question (Figure 7). ANES data (panel a) has a P(VR<0.05) of 0.3. GPT4 (panel b), Llama 405B (panel c), and Llama 70B (panel d) show extreme homogenization, with P(VR<0.05) being close to 1 for all three. Interestingly, Llama8B, which is often less accurate, also tends to be less homogenized, supporting the hypothesized trade-off between accuracy optimization and response diversity.

These findings demonstrate that the homogenization of LLM responses is severe: even when LLMs might provide an answer aligned with the majority opinion of a subgroup, they often do so with a much higher probability than observed in reality, thereby misrepresenting within-group heterogeneity.

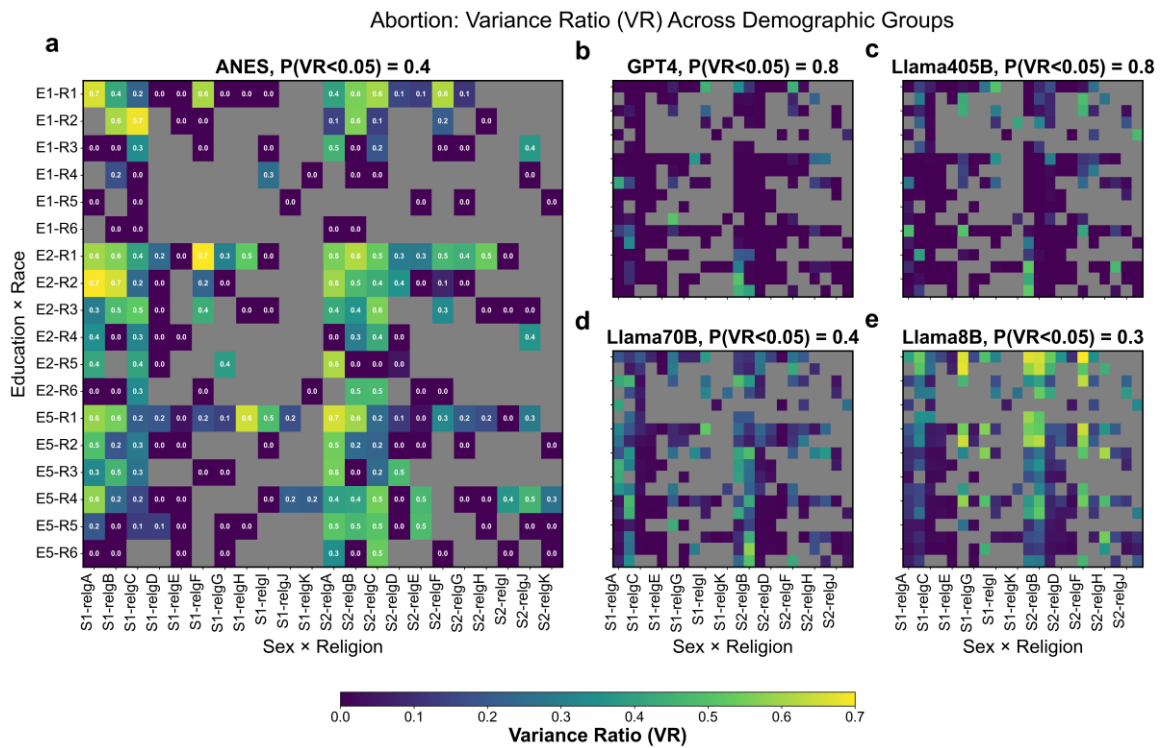

*Figure 6* *Response Variation for Abortion.*

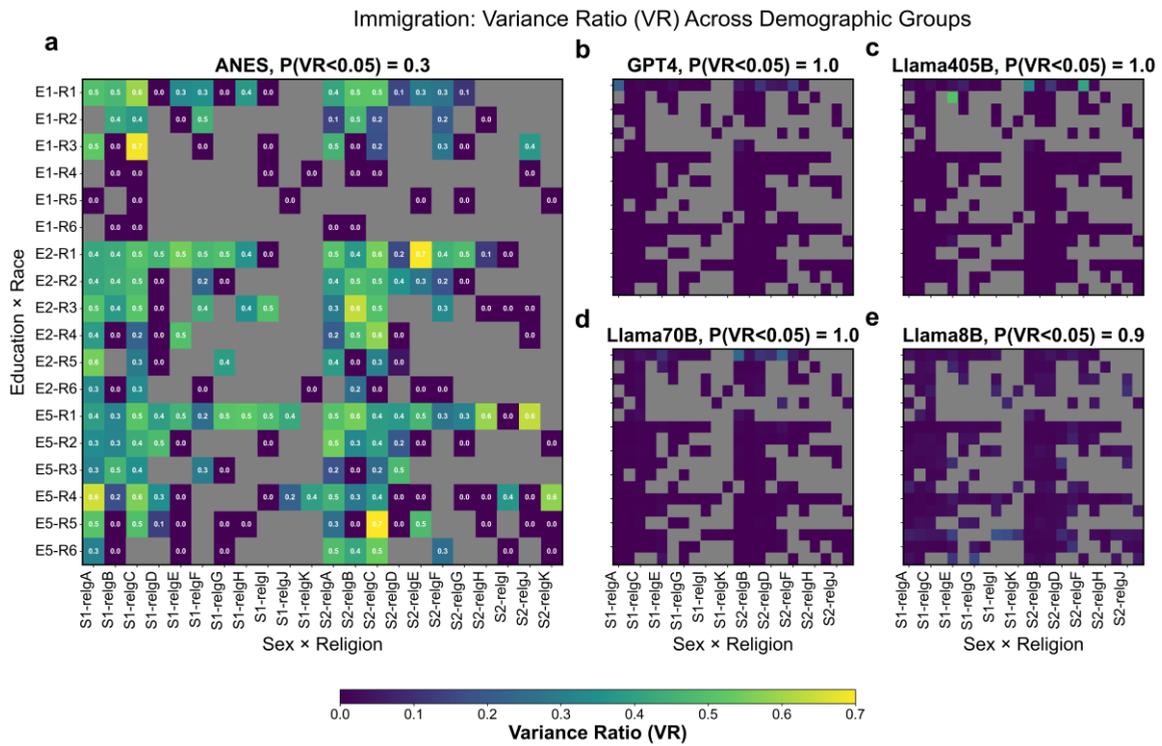

*Figure 7   Response Variation for Immigration.*

**Synthetic variation**

Interpreting these figures in reverse, they also point to a path to get more variation in LLM responses. We call it synthetic variation. Once a persona is defined, we identify what subgroups the persona in question falls into, query the LLM with the same question for all of these subgroups, and then get weighted responses aggregated from these answers. As Figure 3-4 shows, for some questions, this procedure may indeed elevate the variation of the response. For example, when GPT4 is asked what a man would think about the abortion question, instead of giving the direct response which is highly homogenized, increasing the number of variables can give much more varied silicon samples.

Biased sample hypothesis thus is not enough in explaining the lack of answer diversity problem. LLMs will give highly homogenized responses regardless of whether or not the opinions are biased in their innate stock of knowledge, as long as the model chooses to maximize the chance of predicting the user's opinion, as formulated by our accuracy-optimization hypothesis.

**Conclusion and Discussion**

Can LLMs capture the distributions of different opinions in different social groups? This problem poses a unique challenge to LLM because these opinions are often subjective and diverse, without a common ground truth. With nationally representative data, ANES 2020, we find that LLMs do not do well in this task. Our study highlights the potential dangers to uncritically use LLMs to replace human subjects in research. LLMs can produce answers that seem human-like, but they may not be treated as ultimate sources of human ethics or knowledge. The fundamental reason for these limitations lies in how LLMs are trained and what they are optimized for. At their core, LLMs are sophisticated "next token predictors." (Shanahan 2024) They learn statistical patterns from vast amounts of text data to generate plausible sequences of words, rather than developing a genuine understanding of individuals, causality, or human experience.

Our research specifically points to two main problems when LLMs try to represent human opinions.

First, LLM-generated opinions are often inconsistent across different levels of aggregation. This structural inconsistency contrasts with the inherent properties of genuine survey data, where statistics derived from individual responses naturally and reliably aggregate to provide consistent representations at various population strata. This discrepancy, which might seem like a mere technical anomaly, points to an ontological difference between LLMs and the human realities they purport to simulate.

LLM is thus more like an echo of the mass opinion, resembling Heidegger's 'Das Man' (Heidegger 1962) – the anonymous 'They' or the general public. 'Das Man' represents the undifferentiated, everyday understanding and discourse where individuals often passively adopt the views and norms of the crowd, losing their unique and authentic subjectivity. An LLM, optimized to generate probable and widely acceptable text, functions akin to this 'Das Man.' It articulates what 'one says' or what is generally considered plausible within a given context, rather than an opinion rooted in personal conviction or lived experience. When an LLM is prompted with a persona, it attempts to infer the 'Das Man' of that specific demographic – the stereotypical or most commonly expressed views associated with that group in its training data. This is fundamentally different from sampling an actual individual who, even while influenced by social norms, possesses a unique standpoint and internal consistency. Thus, while LLMs can mimic the surface opinion of individuality, they are essentially the opposite of a real person's unique self. The structural inconsistencies we see across different levels are a direct result. The 'Das Man' of a broad group is not just the sum of the 'Das Man' of its parts. Each is a separate construction of general discourse, not a sum of real individual data.

Second, and closely connected to the first point, is the strong homogenization tendency of LLMs. For many personas and different social issues, LLM responses show much less variety than real human opinions. The models consistently overestimate how common the main opinion is within a subgroup. This often happens to an extreme degree. As a result, minority views are ignored or completely erased. This makes the revealed images of social groups look much more like a monolith in LLM, with everyone thinking the same, than they are in reality. Such homogenization is a misrepresentation. It can reinforce stereotypes. It also hides the many different views that truly exist in any human group. Unfortunately, increasing model size doesn't seem to mitigate this problem (Bender et al. 2021).

We propose that these models are typically trained to predict the next token in a sequence, rewarded for minimizing the loss of outputs. This 'accuracy-optimization' strategy, as we have termed it, incentivizes the model to converge on the most statistically probable response for a given prompt, which usually corresponds to the most frequently encountered or dominant viewpoint in its vast training dataset. Even when conditioned on a specific persona, the LLM's primary goal remains to generate a high-probability utterance as that persona, which often translates to reproducing the most stereotypical or commonly cited views associated with that persona. Such stereotypical overestimation thus represents a new dimension of algorithmic culture: where earlier algorithmic systems primarily sorted existing cultural expressions (Striphas 2015), LLMs actively amplify cultural positions, potentially contributing to social misunderstandings through misrepresentation rather than mere classification.

These connected findings – structural inconsistency and opinion homogenization – have several implications for researchers who use silicon samples. If LLMs systematically flatten diversity and produce structurally inconsistent representations of social groups, what kind of 'understanding' are we truly gaining? When people have to use LLMs for synthetic samples (say in marketing research Sarstedt et al. (2024)), the benefits of low cost and fast data generation must be weighed against these limitations. We suggest LLM adopters to at least critically check LLM outputs for structural consistency across different demographic levels and the built-in tendency towards homogeneity. If the problems are found to be severe, then any conclusions drawn should be cautioned. Otherwise, such practices may lead to the design of products and services that cater only to a homogenized, stereotyped 'average' user.

Critically, these two factors should also be examined together. An LLM might seem consistent in its answers for an issue across different subgroups. But this consistency could be misleading. It might happen only because the LLM gives the same dominant answer for all subgroups. If so, aggregation will look consistent, but this 'consistency' hides a deep lack of true representation. In addition to market and consumer research, LLM-based agents (Chuang et al. 2023) may also inherit similar issues when they are used to simulate other social processes.

There are several limitations in this study. First, we primarily focused on specific contentious issues within the socio-political context of the United States, future studies could extend to examine more issues in cross-country contexts. Second, this study should be seen as a critical intervention designed to illuminate fundamental challenges that must be confronted as LLMs become more deeply integrated into our research toolkits. The assumption of direct comparability between LLM outputs and human survey data, which we adopted methodologically from existing studies to demonstrate these discrepancies, is itself a complex area that requires reflection. Future research could also explore alternative frameworks for evaluating LLMs. In addition, we mainly focus on current LLMs that do "fast thinking" rather than emerging AIs that spend more time to respond (such as OpenAI O1, DeepSeek R1), assuming that cultural and political information are already baked in the models. Thus, it would also be interesting to see whether 'slow thinking' chatbot AIs exhibit similar problems.

## References

Argyle, Lisa P., Ethan C. Busby, Nancy Fulda, Joshua R. Gubler, Christopher Rytting, and David Wingate. 2023. "Out of One, Many: Using Language Models to Simulate Human Samples." *Political Analysis* 31(3):337–51.

Ayyamperumal, Suriya Ganesh, and Limin Ge. 2024. "Current State of LLM Risks and AI Guardrails." *arXiv Preprint arXiv:2406.12934*.